\documentclass{article} 
\usepackage{iclr2020_conference,times}

\usepackage{amsmath,amsfonts,bm}

\def\eqref#1{equation~\ref{#1}}

\def\1{\bm{1}}

\DeclareMathAlphabet{\mathsfit}{\encodingdefault}{\sfdefault}{m}{sl}
\SetMathAlphabet{\mathsfit}{bold}{\encodingdefault}{\sfdefault}{bx}{n}

\usepackage{graphicx}
\usepackage{amssymb}
\usepackage{url,hyperref}
\usepackage[nameinlink,noabbrev]{cleveref}
\usepackage{subcaption}

\title{Breast Cancer Detection Using Convolutional Neural Networks}

\author{Simon Hadush Nrea \thanks{Funded by the MU-NMBU}\\
	Department of Computer Science and Engineering\\
	Mekelle Institute of Technology - Mekelle University\\
	Mekelle, Ethiopia \\
	\texttt{\{simon.hadush\}@mu.edu.et}\\
	\AND
	Yaecob Girmay Gezahegn \\
	Mekelle Institute of Technology - Mekelle University\\
	Mekelle, Ethiopia \\
	\texttt{\{yaecob.girmay\}@gmail.com} \\
	\And
	Abiot Sinamo Boltena  \\
	Director General, ICT Sector\\
	FDRE Ministry of Innovation and Technology\\
	Addis Ababa, Ethiopia\\
	\texttt{\{abiotsinamo35\}@gmail.com} \\
	\AND
	Gebrekirstos Hagos \\
	Departhment of Oncology\\
	College of Health Science - Mekelle University\\
	Mekelle, Ethiopia \\
	\texttt{\{gebrekirstoshagos\}@gmail.com} \\	
}

\iclrfinalcopy 
\begin{document}

\maketitle

\begin{abstract}
Breast cancer is prevalent in Ethiopia that accounts 34\% among women cancer patients. The diagnosis technique in Ethiopia is manual which was proven to be tedious, subjective, and challenging. Deep learning techniques are revolutionizing the field of medical image analysis and hence in this study, we proposed Convolutional Neural Networks (CNNs) for breast mass detection so as to minimize the overheads of manual analysis. CNN architecture is designed for the feature extraction stage and adapted both the Region Proposal Network (RPN) and Region of Interest (ROI) portion of the faster R-CNN for the automated breast mass abnormality detection.\\
Our model detects mass region and classifies them into benign or malignant abnormality in mammogram(MG) images at once. For the proposed model, MG images were collected from different hospitals, locally.The images were passed through different preprocessing stages such as gaussian filter, median filter, bilateral filters and extracted the region of the breast from the background of the MG image. The performance of the model on test dataset is found to be: detection accuracy 91.86\%, sensitivity of 94.67\% and AUC-ROC of 92.2\%.\\ 

\end{abstract}

\section{Introduction}
\label{headings}
Breast cancer is one of the most common cancer and cause of death among women globally\cite{trimble2017breast, world2006guidelines, mutebi2014stigma}. According to the global cancer statics, the number of new cases in 2018 was estimated to be 18,078,957 and deaths 9,555,027 (52.85\%) globally \cite{bray2018global}. The cases of breast cancer amounts to 2,088,849 (11.55\%) and the deaths is estimated to be 626,679 (6.56\%). 60\% of the deaths occur in low income developing countries like Ethiopia \cite{bray2018global,hadgu2018breast,vanderpuye2019cancer, stefan2015cancer}.Studies show that cancer diagnosis in Ethiopia is time consuming, in some cases it takes three to six months in order to report a given subject is positive or negative. During this time interval the disease could reach uncontrollable stage before the cancer is deemed positive and this may lead to lower survival rate  \cite{vanderpuye2019cancer, dibisa2019breast}. Mammography is the golden standard imaging modality used to detect breast abnormalities at an early stage\cite{sree2011breast} and hence, micro calcification and masses are the earliest signs of breast cancer which can only be detected using imaging modality. The abnormalities may be benign or malignant depending on the invasive stage of the breast abnormality. Detection of masses in breast tissue is more challenging compared to the detection of micro calcification \cite{hagos2018improving}. Early detection of breast cancer has shown a reduction of mortality rate between  38\% to 48\% \cite{broeders2012impact}. However, manual mammogram analysis and interpretation leads to 10\% - 30\% misdiagnosis error rates \cite{oeffinger2015breast, kerlikowske2000performance}.\\
Lack of early detection leads to thousands of women go through painful, lower survival rate and scar inducing surgeries. To mitigate this and similar challenges many studies have been undertaken using both conventional machine learning and deep learning based methods.\\
Recently, due to huge amount of data, high computational power of Graphical Processing Units (GPUs), deep Learning has shown promising success in natural language processing \cite{iyyer2015deep}, object detection and recognition \cite{goodfellow2016deep} and medical image analysis \cite{shen2017deep, greenspan2016guest, ben2017domain}. Deep learning based methods are sensitive to image acquisition setting, scanner types and the image preprocessing applied. Moreover, the studies by \cite{martin2000genetic,world2006guidelines} showed that the evolution of breast cancer can be shaped by race, geographical location, and other risk factors. In this work, we proposed convolutional neural network(CNN) based breast mass detection approach to simultaneously localize and classify the mass into either benign or malignant abnormality. To train, validate and test the method, datasets were collected from different sites.\\
Overall, this work has the following contributions:
\begin{itemize}
	\item Here in Ethiopia, we had experienced a hard time in searching and collecting the required dataset. Hence, we have collected the dataset that will be useful as a starting point for researchers who will be working in the area. 
	\item An automated breast mass abnormality detection model was developed that detects and localizes mass regions and classifies them into benign or malignant in MG images.
\end{itemize}
\section{Methodology}
\subsection{\label{ethical} Ethical statement and Confidentiality}
Ethical clearance was given from College of Health Sciences - Mekelle University, by the Institutional Review Board (IRB) and the ethical review committee of each respective hospitals where data was collected. Profiles of the patients were removed and anonymised in order not to be identifiable in any way, and kept the confidentiality of the gathered data with great care.
\subsection{Dataset}
\begin{figure}[h]
	\begin{center}
	\begin{subfigure}[b]{0.3\textwidth}
		\includegraphics[width=2.5cm, height=3.5cm]{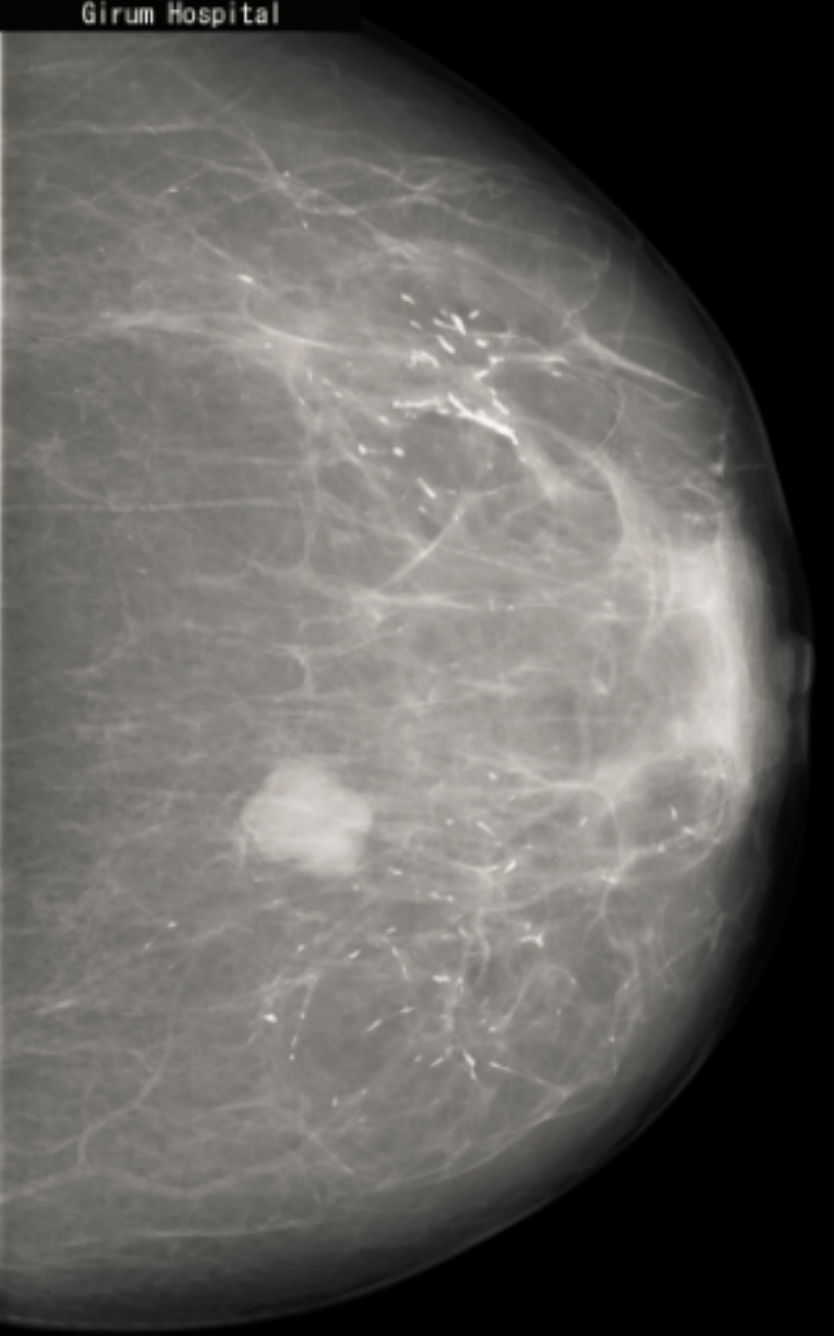}
		\caption{}
		\label{GR}
	\end{subfigure}
	\begin{subfigure}[b]{0.3\textwidth}
		\includegraphics[width=2.5cm, height=3.5cm]{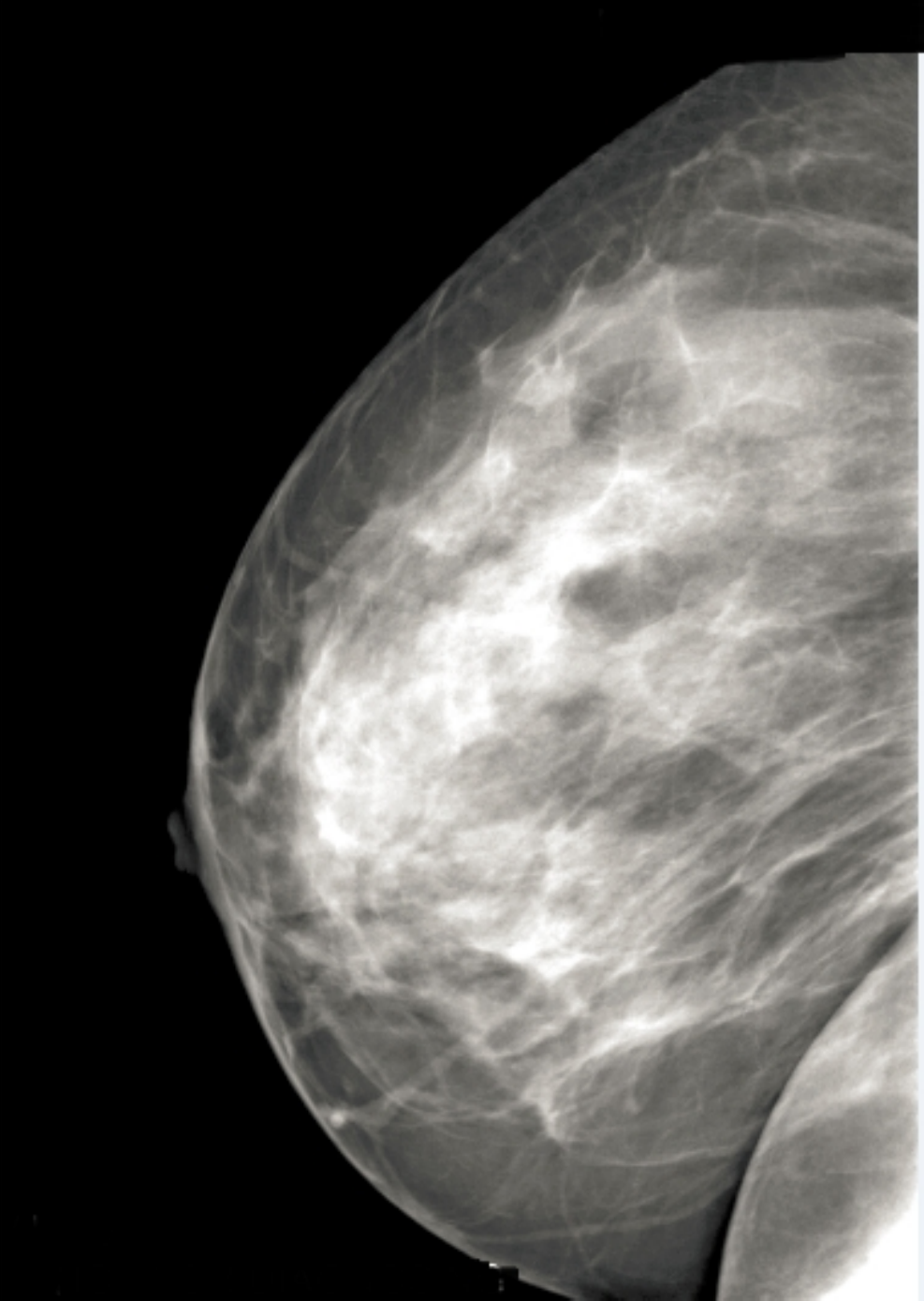}
		\caption{}
		\label{pdiag}
	\end{subfigure}
	\begin{subfigure}[b]{0.3\textwidth}
		\includegraphics[width=2.5cm,height=3.5cm]{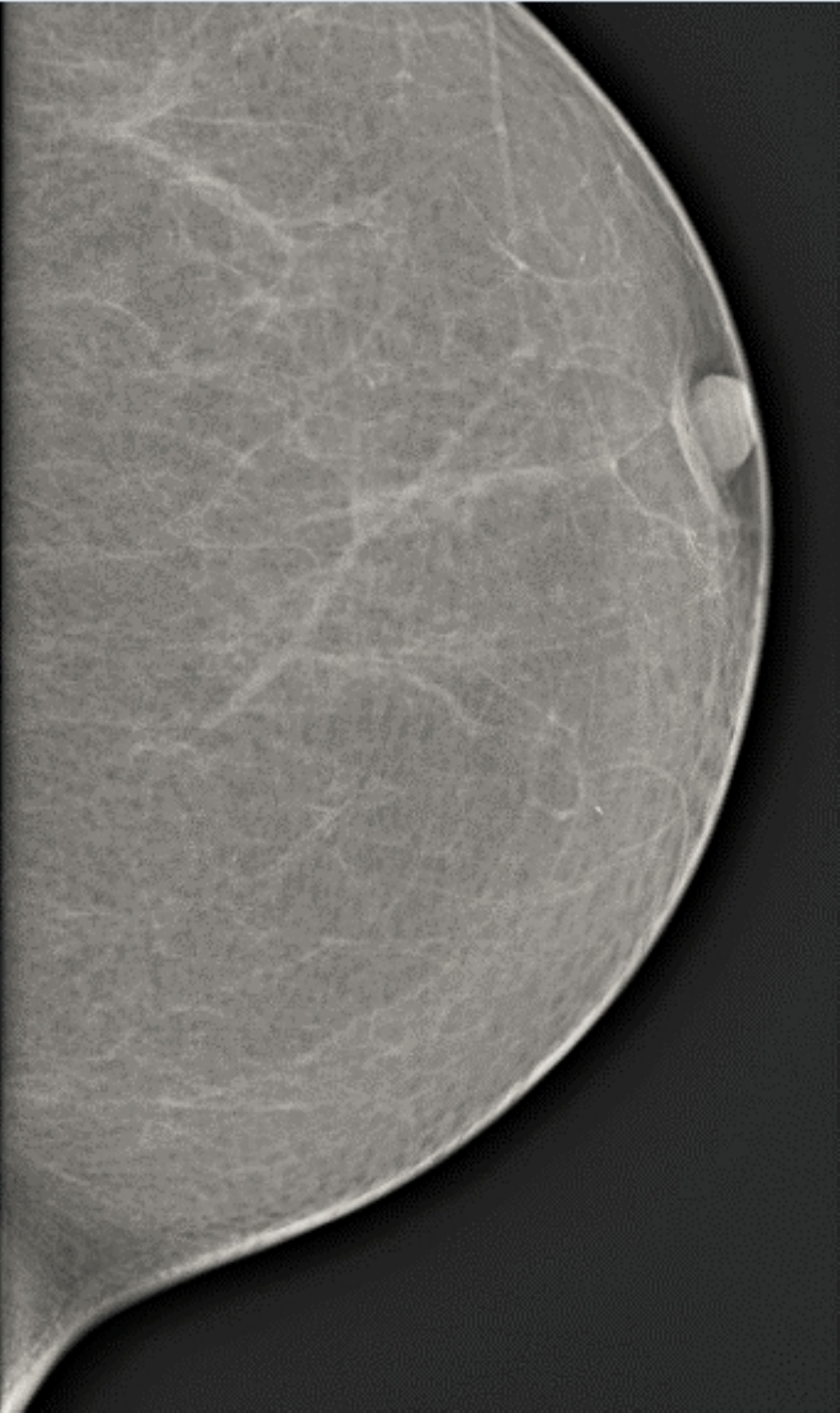}
		\caption{}
		\label{stGRH}
	\end{subfigure}	
	\end{center}
		\caption{Sample of MG images during acquisition stage. The mammogram images are from different mammography x-ray sources: (a) from Girum Hospital, (b) from Pioneer Diagnostic Center, and (c) from St. Gebriel Hopital.}
	\label{sample_MG_images}
\end{figure}

The dataset was mainly collected from St.Gebriel Hospital, Grum Hospital, Betezatha Hospital, Korean Hospital, Kadisco Hospital and Pioneer Diagnostic. The MG images were collected with their document reports that show the screening and diagnosis results of the patients. The documents report results were based on the pathology confirmation and Breast Imaging-Reporting and Data System (BI-RADS). More than 5000 x-ray mammogram images that were diagnosed between 2016 and 2018 as shown in \cref{list_of_MG_images} were collected and some of the samples are shown in \cref{sample_MG_images}. This work considered only the mass abnormality from the collected MG images, that is 1588  full mammogram images which have mass abnormality and annotated by professional radiologists using the labelMe \cite{russell2008labelme} annotation tool. The dataset was divided into training (80\%), validation (10\%), and testing (10\%).  
\begin{table}[t!]\centering
	\def\arraystretch{1}%
	\vspace{0.4cm}
	\caption{List of total collected MG images from different hospitals and the number of MG images consisting of mass abnormalities for both benign and malignant. In this work, mass abnormality was considered.}
	\label{list_of_MG_images}
	\begin{tabular}{ |p{3cm}|p{2.2cm}|p{2cm}|p{2cm}|p{2cm}|p{2cm}|  }
		\hline
		Dataset Source & Total patient cases ( total MG images) &  Total number of MG images with mass abnormality &  Training & Validation& Testing \\
		\hline
		St. Gebriel Hospital & 580(2224) & 800 & 640& 80& 80 \\
		\hline
		Grum Hospital & 450(1684) & 400 & 320& 40& 40 \\
		\hline
		Korea Hospital & 280(1024) & 210 & 168 & 21& 21 \\
		\hline
		Betezatha  Hospital & 340(1270) & 138 & 110 & 14& 14 \\
		\hline
		Kadisko Hospital & 20(70) & 40 & 32 & 4& 4 \\
		\hline
		Pioneer Diagnostic Center & 20(68) & 30 & 24 & 3& 3 \\
		\hline
	\end{tabular}
\end{table}
\subsection{Method}
\begin{figure}[h]
	\begin{center}
	\includegraphics[scale=0.8]{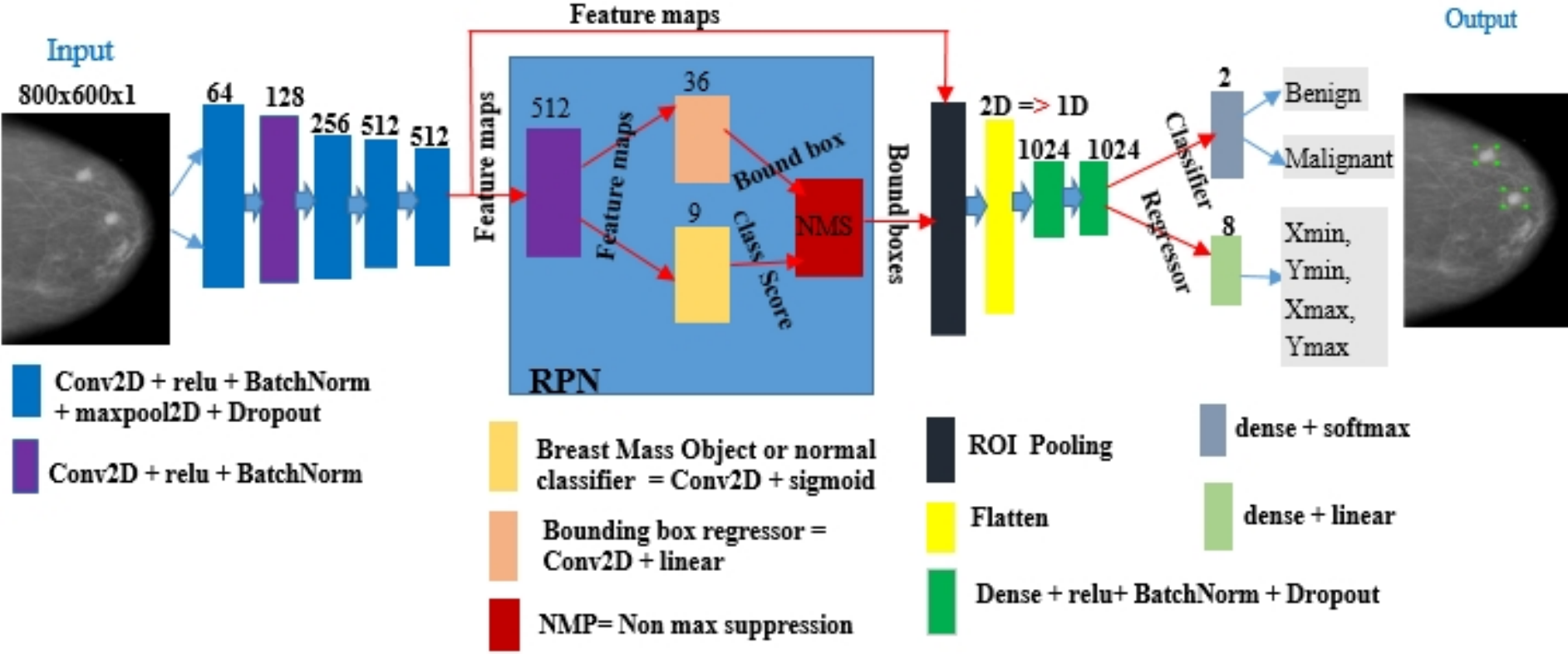}	
	\end{center}
	\caption{ Structure of Breast Cancer Detection architecture flow work. The detection portion of RPN and ROI pooling are adapted from the Faster R-CNN \protect\cite{ren2015faster} }
    \label{detection_model}
\end{figure}

We developed CNN based breast mass abnormality detection model that detects automated region of interest of mass abnormality and classify them into benign or malignant in MG images. We used preprocessing and augumentation for the 116 full MG images taken from INbreast \cite{moreira2012inbreast} and  1380 full MG images taken from CBIS-DDSM \cite{lee2017curated} so as to have initial weights for training our model and the local collected dataset. Overall, our proposed method has the following steps:\\
\begin{itemize}
	\item Data collection: the dataset described in \cref{list_of_MG_images} was collected from different hospitals in Ethiopia.\\
	\item MG image preprocessing: To enhance the quality of the data and prepare it in a suitable way for deep learning training the data was preprocessed. To remove noise in the images gaussian filtering, median and bilateral filtering were applied. Later the images were enhanced using contrast-limited adaptive histogram equalization (CLAHE) and then followed by morphological operation and OTSU’s thresholding to extract the breast region from the background and to remove non-portion of breast region from the MG such as artifacts, labels, patient profiles and others.\\
	\item Model training: \cref{detection_model} shows our proposed  breast cancer detection pipeline. The feature extraction section has a series of five convolutional layers with (64, 128, 256, 512,512 ) number of filters for each convolution layer, respectively. Each convolution layer is followed by Relu activation layer, batch normalization, maxpooling layer and dropout except the second layer which has neither dropout nor maxpooling. Convolution was performed with a kernel filter of (3,3) , stride of (1,1) and same padding. The max pooling was performed with stride of (2,2) and kernel filter of (2,2). Furthermore, by adapting the anchor's bound box scales, ratios of the RPN and maxpooling of the ROI Pooling portion of the Faster R-CNN \cite{ren2015faster}, it was employed for the detection of mass abnormalities. We used 9 anchors with 32 $\times$ 32, 64 $\times$ 64 and 128 $\times$ 128 pixels of anchor box scales  and [1, 1], [1./sqrt(2), 2./sqrt(2)] and [2./sqrt(2), 1./sqrt(2)] of anchors box ratios, and (5,5) ROI maxpooling size. What is more, 0.9 momentum, 500 epochs, 0.00001 learning rate, Adam for the RPN and Stochastic Gradient Descent(SGD) for the whole model as optimizers were used.
	The proposed model was implemented with Python and Keras, where Tensorflow was used as a backend.
\end{itemize}
\section{Result and Discussion}
\begin{figure}[h]
	\begin{center}
	\begin{subfigure}[b]{0.4\textwidth}
		\includegraphics[width=5cm, height=4cm ]{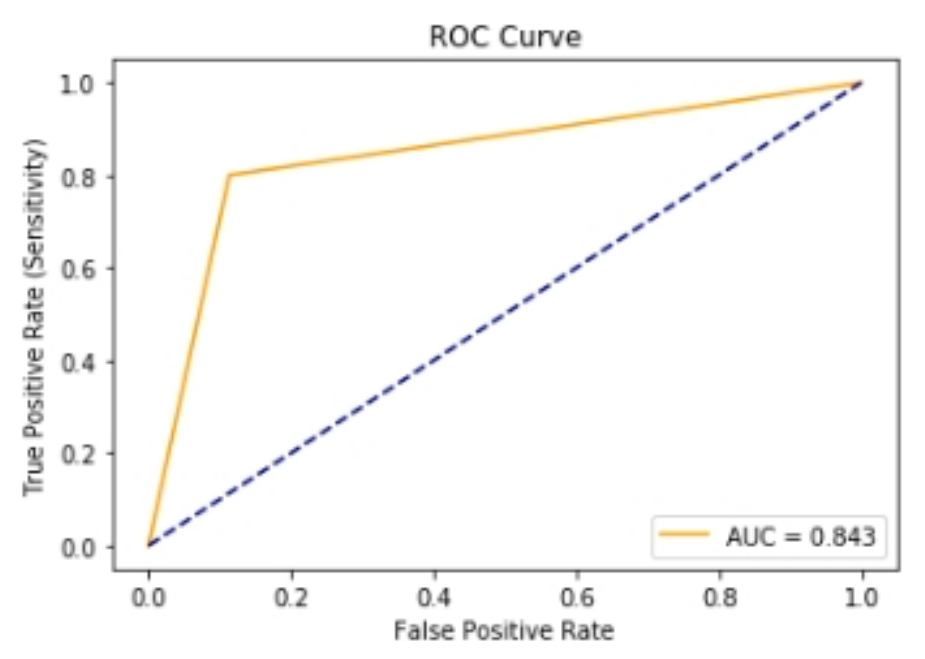}
		\caption{}
		\label{vgg_frcnn_ROC_cure}
	\end{subfigure}
	\begin{subfigure}[b]{0.4\textwidth}
		\includegraphics[width=5cm, height=4cm ]{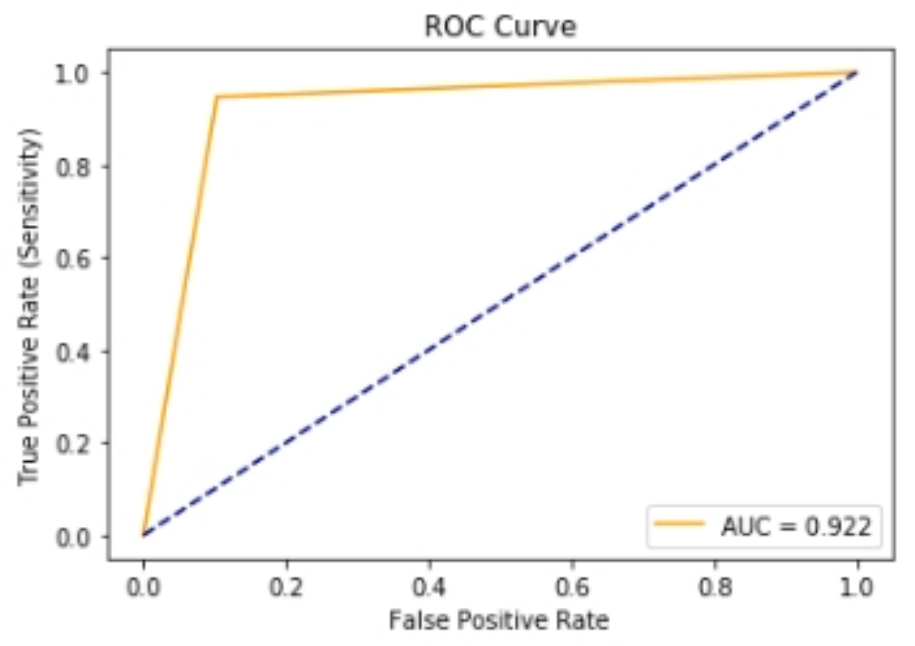}
		\caption{}
		\label{proposed_ROC_cure}
	\end{subfigure}
   \end{center}
	\caption{ROC curves of (a) VGG based faster R-CNN model  and (b) the proposed model }
	\label{}
\end{figure}

\begin{figure}[h]
	\begin{center}
		\begin{subfigure}[b]{0.3\textwidth}
		\includegraphics[width=2.5cm, height=3cm ]{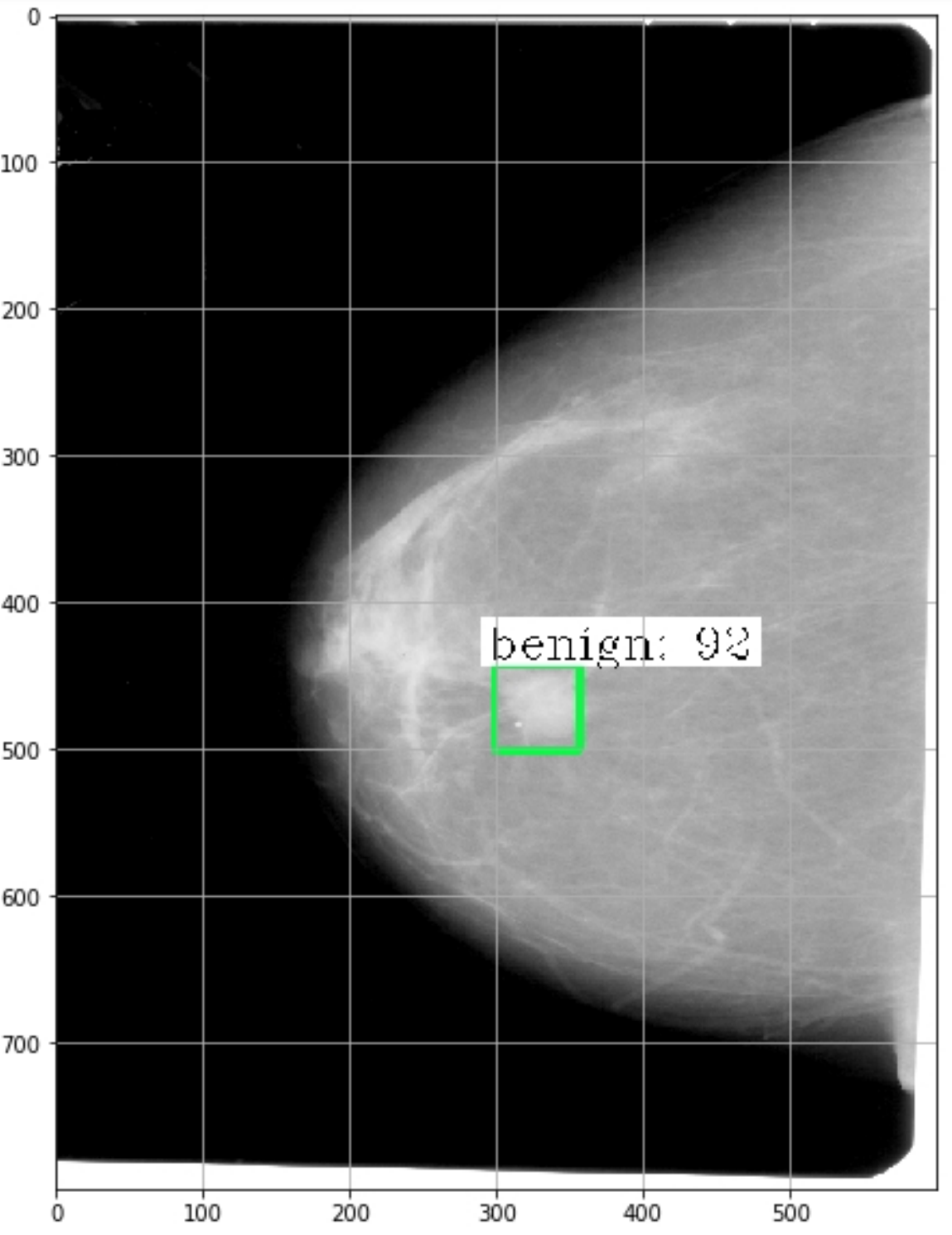}
		\caption{}
		\label{}
	\end{subfigure}
	\begin{subfigure}[b]{0.3\textwidth}
		\includegraphics[width=2.5cm, height=3cm ]{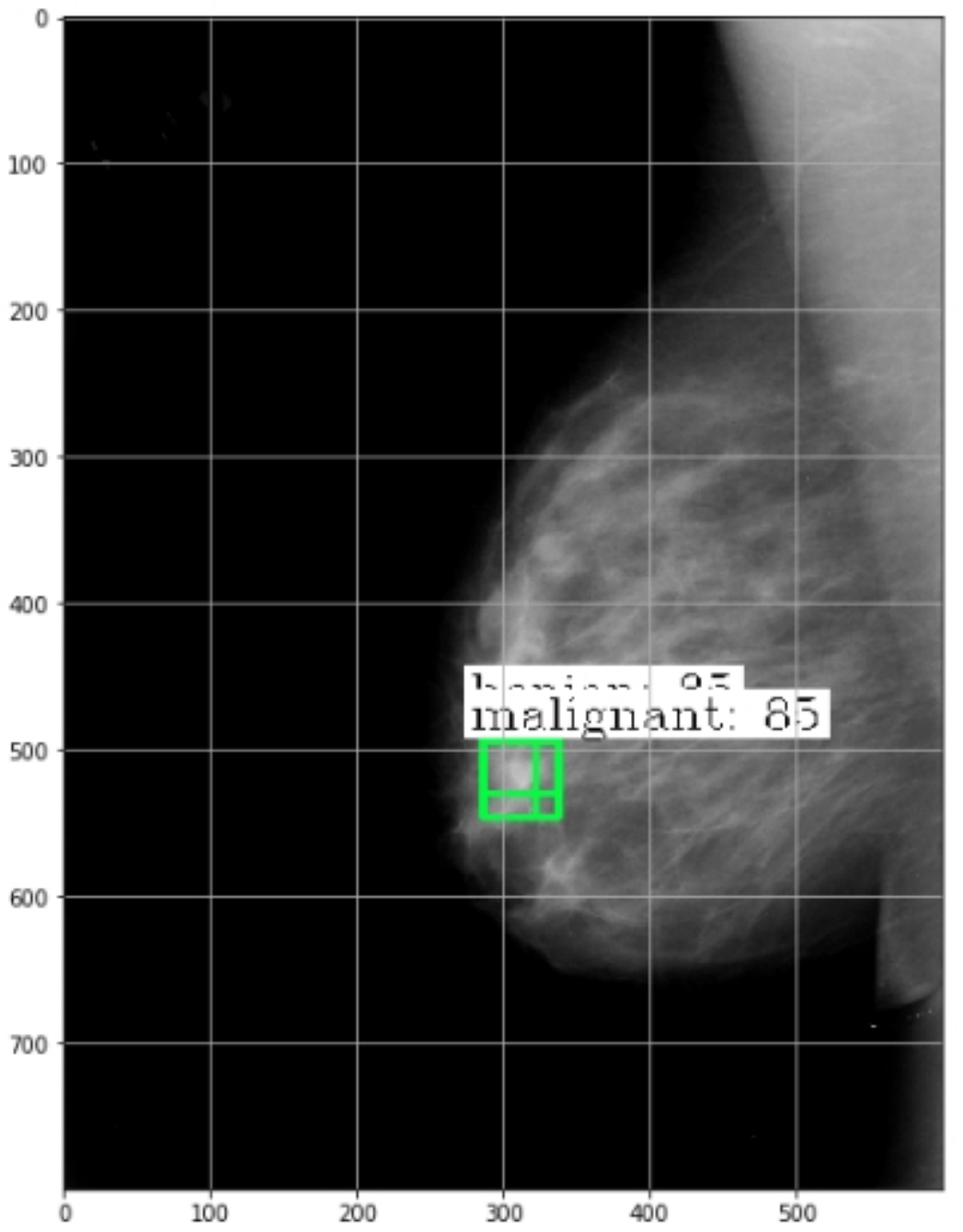}
		\caption{}
		\label{}
	\end{subfigure}
	\begin{subfigure}[b]{0.3\textwidth}
		\includegraphics[width=2.5cm, height=3cm ]{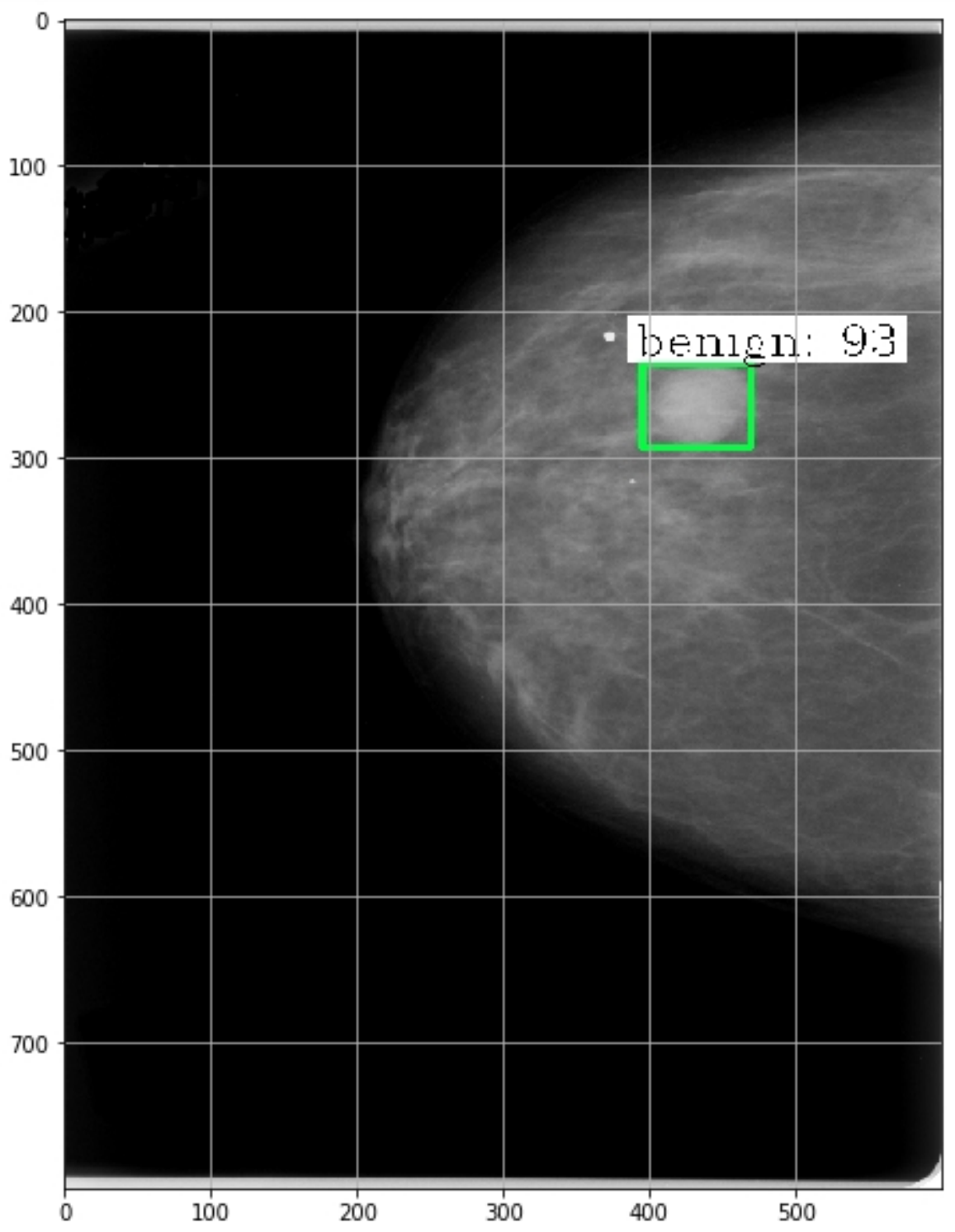}
		\caption{}
		\label{}
	\end{subfigure}
	\begin{subfigure}[b]{0.3\textwidth}
		\includegraphics[width=2.5cm, height=3cm ]{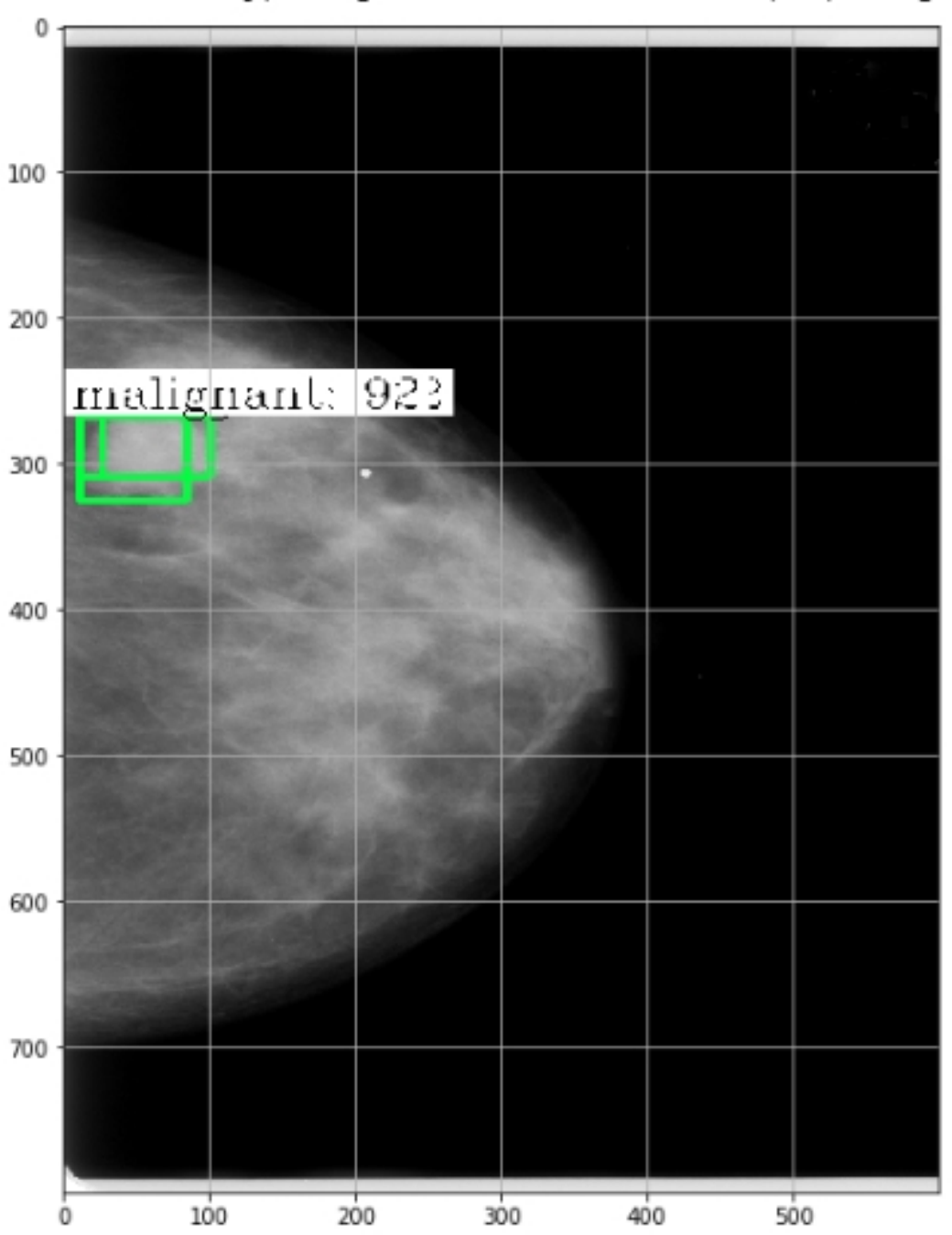}
		\caption{}
		\label{}
	\end{subfigure}
	\begin{subfigure}[b]{0.3\textwidth}
		\includegraphics[width=2.5cm, height=3cm]{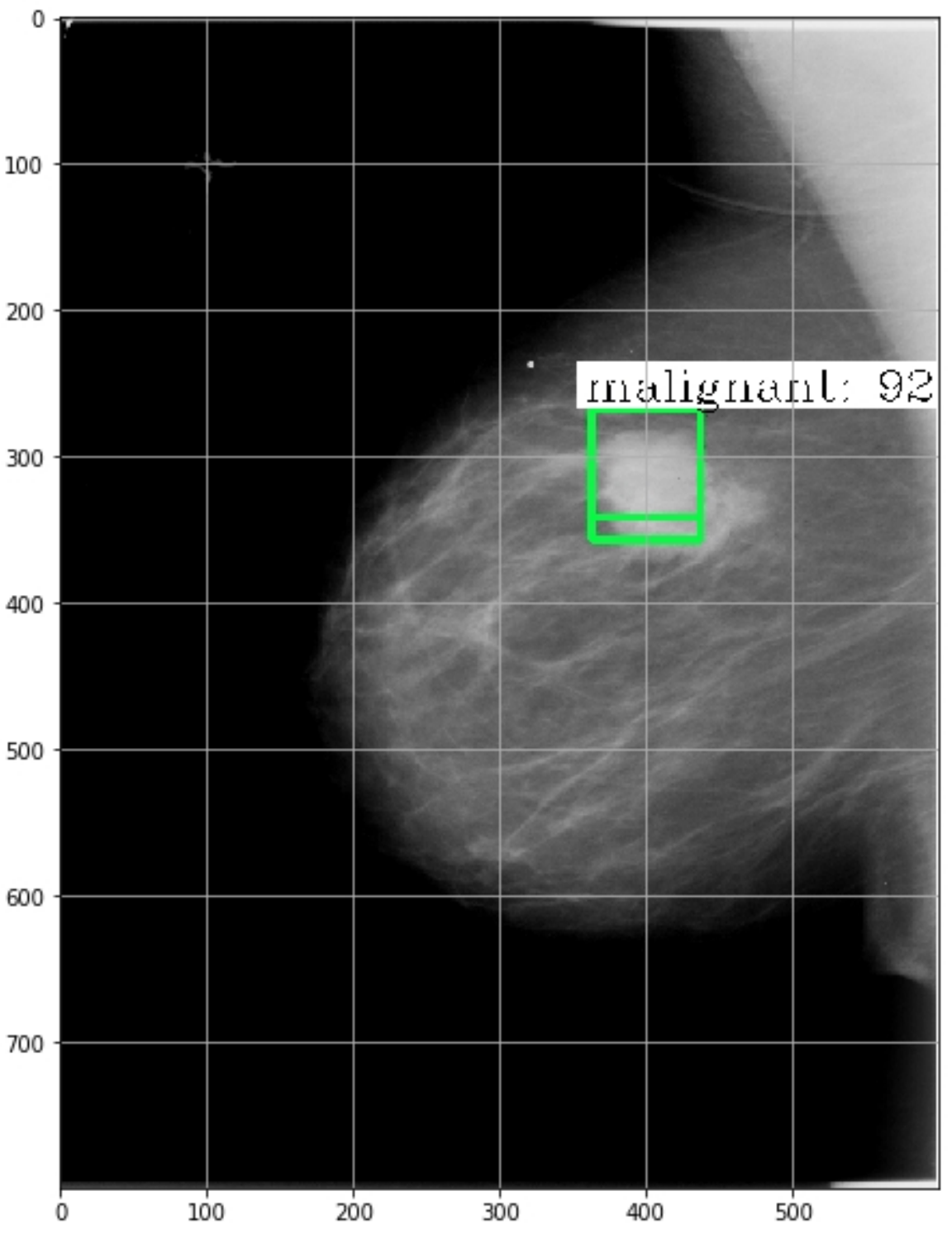}
		\caption{}
		\label{}
	\end{subfigure}
	\begin{subfigure}[b]{0.3\textwidth}
		\includegraphics[width=2.5cm, height=3cm]{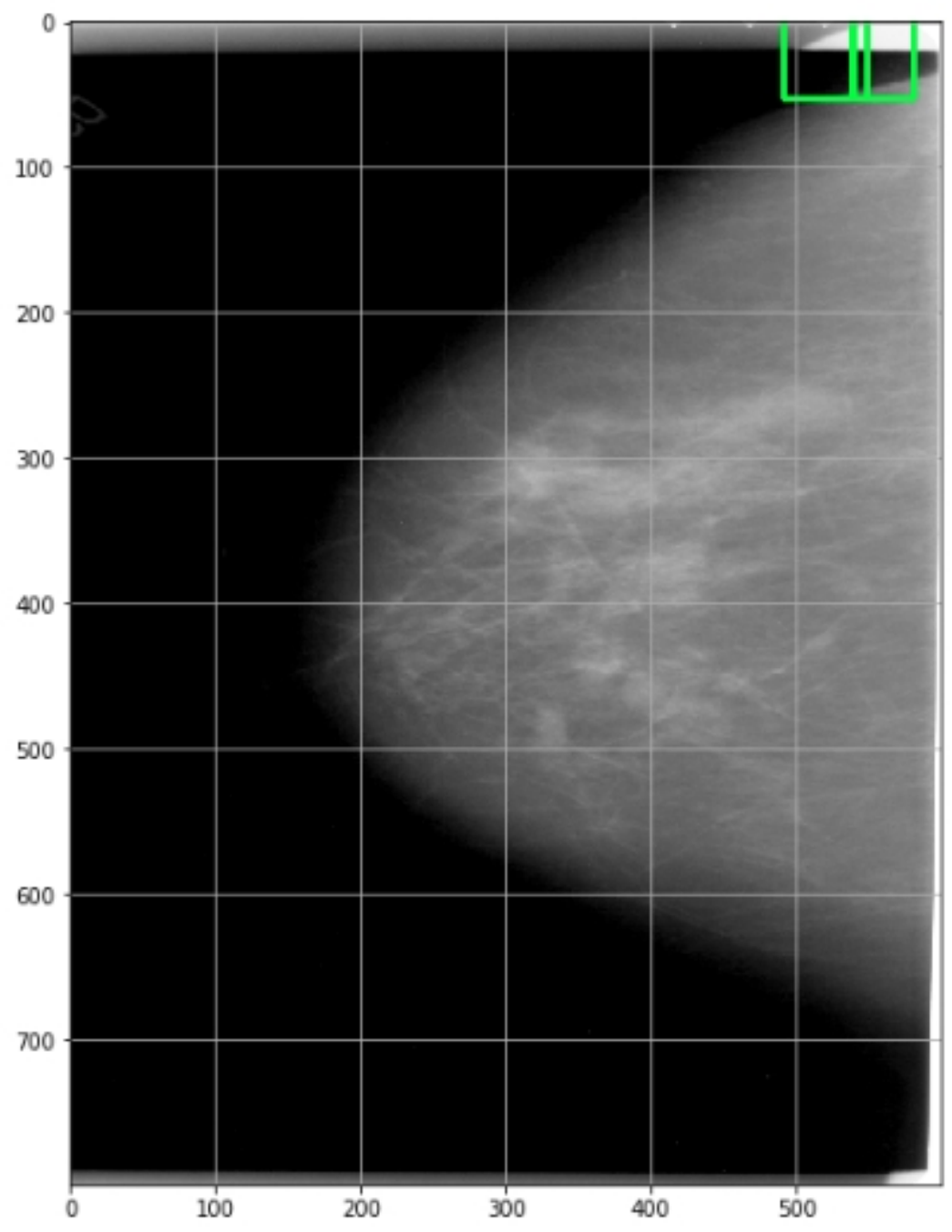}
		\caption{}
		\label{}
	\end{subfigure}
	\end{center}
  \caption{Sample of mass abnormalities (a) to (b) detected by VGG based faster R-CNN model, (c) to (e) detected and (f) miss-detected by the proposed model.}
\label{detected_MG_images}
 \end{figure}

This paper describes a CNN based approach for detecting mass regions and classifies them into benign or malignant. Here, mass abnormality detection, localization and classification in to benign or malignant in one process in local multi-center MG data-set is investigated. It is difficult to directly compare our detection results with previous works, locally. Hence, we have trained, validated and tested VGG based faster R-CNN \cite{ren2015faster} architecture in order to compare with our model's performance by using the dataset that was collected. Out of the total collected images, 1588 full MG images that contain mass abnormalities were selected and then annotated by professional radiologists using the labelMe\cite{russell2008labelme} annotation tool. In the 1588 MG images there are 1683 number of breast mass abnormalities. The dataset was randomly split into 80\% for training, 10\% for validation and 10\% for testing. We performed the same pipeline preprocessing for INbreast \cite{moreira2012inbreast}, CBIS-DDIS \cite{lee2017curated} and collected local MG data-sets for the proposed model and VGG based faster R-CNN. In the preprocessing stage: different imaging formats such as DICOM medical image format to .png image format were converted, noise was removed, breast region was extracted from the background, patients' information were removed, artifacts and other unwanted objects were cleansed. Gaussian, medium and belaterial filters with 3$\times$3 and 5$\times$5 sizes each were used for noise removal and evaluated the denosied results using MSE. Out of the two filter sizes considered the one with 3$\times$3 size was finally used. Additionally, CLAHE was used to enhance the denoised MG images and after that the breast region was extracted and unwanted artifacts were removed using OTSU and morphological operations.\\ 
Four different threshold values such as T=1 (100\%), 0.75 (75\%), 0.5 (50\%) and 0.25 (25\%) were used in the experiments for the Intersection over Union (IoU) overlapping the ground truth bound box and predictive bound box in the RPN. The IoU overlap are summarized in \cref{IoU_vgg_results} for the VGG based faster R-CNN detection model and in \cref{IoU_our_model_results} for the proposed detection model. In both models the IoU score improves as the threshold value decreases. Since the breast mass abnormalities are very small compared with the natural image objects such as person face, car and other objects in imagenet\cite{deng2009imagenet}. So, the final threshold value is set to T=0.25 which is more suitable to detect almost all mass abnormalities.
\begin{table}[t!]\centering
	\def\arraystretch{1}%
	\vspace{0.4cm}
	\caption{VGG based faster R-CNN detection model: IoU over all results with varying overlap percentage thresholds for the test MG dataset}
	\label{IoU_vgg_results}
	\begin{tabular}{ |p{4cm}|p{1.8cm}|p{1.8cm}|p{1.8cm}|p{1.8cm}|  }
		\hline
		Overlap threshold T & 100\% & 75\% & 50\%& 25\% \\
		\hline
		IoU & 28.46\% & 49.5\% & 69.25\%& 88.76\% \\
		\hline
	\end{tabular}
\end{table}
\begin{table}[t!]\centering
	\def\arraystretch{1}%
	\vspace{0.4cm}
	\caption{Our proposed detection model: IoU overall results with varying overlap percentage thresholds for the test MG dataset}
	\label{IoU_our_model_results}
	\begin{tabular}{ |p{4cm}|p{1.8cm}|p{1.8cm}|p{1.8cm}|p{1.8cm}|  }
		\hline
		Overlap threshold T & 100\% & 75\% & 50\%& 25\% \\
		\hline
		IoU & 36.26\% & 58.75\% & 78.34\%& 94.2\% \\
		\hline
	\end{tabular}
\end{table}
The classification result is summarized as shown in \cref{vgg_class_result_table} and \cref{pr_class_result_table}, and  ROC curves are shown in \cref{vgg_frcnn_ROC_cure} and \cref{proposed_ROC_cure} for the VGG based Faster R-CNN and the proposed model, respectively. The result for the Area Under the Curve (AUC) for the proposed model is (AUC=92.2 \%) in taking the right decision taken during the classification than the AUC of the VGG based faster R-CNN detection model ( AUC= 84.3\%).
Overall, the proposed model performs well than VGG based faster R-CNN. The model outperformed because the hyperparameters of the designed feature extractor (base network) and the adapted RPN and ROI pooling detection portion were tweaked. Besides, the model prevents overfitting and internal covariance shift due to  dropout and batch normalization layers in both the base network and RPN in each after the non-linear activation layer’s (Relu), respectively.
It was observed that there are feature similarity between the benign and malignant. The reason is that data was collected from different hospitals under different lighting conditions, scanned using different mammography devices, and other factors. Hence, VGG based Faster R-CNN detection model misclassified  26 and  the proposed model misclassified 14. 
\begin{table}[t!]\centering
	\def\arraystretch{1}%
	\vspace{0.4cm}
	\caption{VGG based faster R-CNN detection model performance: over all breast mass abnormality classification results of MG images.}
	\label{vgg_class_result_table}
	\begin{tabular}{ |p{2.2cm}|p{2.2cm}|p{2.2cm}|p{2.2cm}|p{2.2cm}|p{2.2cm}|  }
		\hline
		Evaluation Metrics & Accuracy  & Sensitivity (Recall) & Specificity & Precision & AUC-ROC \\
		\hline
		Results in percentage & 84.88\% & 80\% & 88.65\%& 84.5\% & 84.3\% \\
		\hline
	\end{tabular}
\end{table}
\begin{table}[t!]\centering
	\def\arraystretch{1}%
	\vspace{0.4cm}
	\caption{The proposed detection model performance: over all breast mass abnormality classification results in MG images.}
	\label{pr_class_result_table}
	\begin{tabular}{ |p{2.2cm}|p{2.2cm}|p{2.2cm}|p{2.2cm}|p{2.2cm}|p{2.2cm}|  }
		\hline
		Evaluation Metrics & Accuracy  & Sensitivity (Recall) & Specificity & Precision & AUC-ROC \\
		\hline
		Results in percentage & 91.86\% & 94.67\% & 89.69\%& 87.65 \% & 92.2\% \\
		\hline
	\end{tabular}
\end{table}
Some of the mass abnormality detected from the test dataset by the VGG based Faster R-CNN and the proposed model are shown in \cref{detected_MG_images} (a) and (b), and (c) to (e), respectively.  \cref{detected_MG_images} (f) is one of the miss-detected in the MG image by the model. The green bounding boxes in the figures  are showing the detected mass abnormalities. Each detected mass abnormalities contains bounding box with green color, class name and confidence score.
\section{Conclusion}
The paper presents CNN based breast cancer detection model that detects mass abnormality and classify them in to benign and malignant in MG. A CNN architecture for feature extraction was designed as shown in \cref{detection_model} and RPN and ROI pooling portion from the faster R-CNN \cite{ren2015faster} for the breast mass abnormality detection was adopted.\\
The model was trained and evaluated via the MG images. The proposed model achieved detection accuracy of up to 91.86\% , sensitivity of 94.67\%  and AUC-ROC of 92.2\%.\\
Based on the investigation and findings of the study, the following recommendations are forwarded for further research works:
\begin{itemize}
	\item We have only considered breast mass abnormality. So, the study can be extended to include macro calcification abnormalities that is not considered here.
	\item By increasing the amount of annotated MG dataset, the performance of the model can be enhanced.
\end{itemize}
\subsection*{Acknowledgments}
We would like to thank Mekelle University and MU-NMBU for their financial support. We would also like to thank St. Gebriel Hospital, Grum Hospital, Kadisco Hospital, Korea Hospital, Pioneer Diagnostic Center and Betethata Hospital for their cooperation and generosity in providing the datasets.
\bibliography{iclr2020_conference}
\bibliographystyle{iclr2020_conference}
\end{document}